\documentclass[journal]{IEEEtran}
\usepackage{times}
\usepackage{epsfig}
\usepackage[belowskip=-10pt,aboveskip=5pt]{caption}
\usepackage{graphicx}
\usepackage{amsmath}
\usepackage{booktabs}
\usepackage{bm}
\usepackage{nomencl}
\makenomenclature
\usepackage{boldline,multirow} 
\usepackage[caption=false,font=footnotesize]{subfig}
\usepackage[vlined, ruled,linesnumbered]{algorithm2e}
\usepackage[shortlabels]{enumitem}
\usepackage{makecell}
\usepackage[normalem]{ulem}
\useunder{\uline}{\ul}{}
\usepackage{floatrow}
\newfloatcommand{capbtabbox}{table}[][\FBwidth]

\usepackage[capitalise]{cleveref}
 \usepackage{amssymb}

\ifCLASSINFOpdf

\else

\fi

\begin{document}

\title{IDE-Net: Interactive Driving Event and Pattern Extraction from Human Data}

 \author{
 Xiaosong Jia, 
 Liting Sun, 
 Masayoshi Tomizuka, 
 and Wei Zhan
 \thanks{X. Jia, L. Sun, M. Tomizuka and W. Zhan are with University of California, Berkeley. \{jiaxiaosong, litingsun, wzhan, tomizuka\}@berkeley.edu}
}

\maketitle

\begin{abstract}
Autonomous vehicles (AVs) need to share the road with multiple, heterogeneous road users in a variety of driving scenarios. It is overwhelming and unnecessary to carefully interact with all observed agents, and AVs need to determine \textit{whether} and \textit{when} to interact with each surrounding agent. In order to facilitate the design and testing of prediction and planning modules of AVs, in-depth understanding of interactive behavior is expected with proper representation, and events in behavior data need to be extracted and categorized automatically. Answers to \textit{what} are the essential patterns of interactions are also crucial for these motivations in addition to answering \textit{whether} and \textit{when}.
Thus, learning to extract interactive driving events and patterns from human data for tackling the \textit{whether-when-what} tasks is of critical importance for AVs. 
There is, however, no clear definition and taxonomy of interactive behavior, and most of the existing works are based on either manual labelling or hand-crafted rules and features. 
In this paper, we propose the \textbf{I}nteractive \textbf{D}riving event and pattern \textbf{E}xtraction \textbf{Net}work (IDE-Net), which is a deep learning framework to automatically extract interaction events and patterns directly from vehicle trajectories. In IDE-Net, we leverage the power of multi-task learning and proposed three auxiliary tasks to assist the pattern extraction in an unsupervised fashion. We also design a unique spatial-temporal block to encode the trajectory data. Experimental results on the INTERACTION dataset verified the effectiveness of such designs in terms of better generalizability and effective pattern extraction. We find three interpretable patterns of interactions, bringing insights for driver behavior representation, modeling and comprehension. Both objective and subjective evaluation metrics are adopted in our analysis of the learned patterns. 
\end{abstract}


\IEEEpeerreviewmaketitle

\section{Introduction}
\subsection{Motivation}
\IEEEPARstart{I}{n}  various driving scenarios, autonomous vehicles need to interact with other road users such as vehicles, pedestrians and cyclists. It is computationally overwhelming and unnecessary for autonomous vehicles to pay equal attention to all detected entities simultaneously. Therefore, it is desired if we can identify \textit{whether} each object may potentially interact with the ego vehicle, and \textit{when} would be the starting and ending points of the interaction period, so that limited resources can be better allocated to tackle high-priority entities timely.

Meanwhile, predicting complex human driving behavior and designing human-like behaviors for autonomous vehicles are always challenging due to our insufficient understanding of interactive human behavior. Motions of interactive agents are intrinsically high dimensional. Desirable answers to \textit{what} essential representation (with patterns) of such motions can be extracted or learned will significantly facilitate prediction and behavior modeling \cite{brown2020modeling,sun2019interpretable}, as well as imitation \cite{codevilla2018end,rhinehart_deep_2019}, decision and planning \cite{Zeng_2019_CVPR} in terms of performances and computational efficiency with in-depth comprehension in a much lower dimensional space.

Moreover, with more and more trajectory data obtained from bird's-eye view \cite{interactiondataset} and ego-vehicle perspective \cite{Sun_2020_CVPR}, it is desired to learn to automatically extract interactive driving events (\textit{whether} and \textit{when}) and properly tag/categorize them (\textit{what}) from large amounts of trajectory data in order to efficiently train and test behavior-related algorithms, as well as to generate scenarios and behavior for testing. 

\subsection{Related Works}
\subsubsection{Agent prioritization and scenario representation}
In order to rank the priority of surrounding entities of the ego vehicle in a driving scene, \cite{refaat2019agent} constructed a convolutional neural network to learn the ranking via supervised learning. Prioritization can also be implicitly learned via graph neural networks in prediction tasks \cite{hu2020scenario}. However, the answers to \textit{whether}, \textit{when} and \textit{what} for extracting interactions were not explicitly provided in these works.
A comprehensive scene and motion representation framework was proposed in \cite{hu2020scenario} to construct semantic graph for interactive vehicle prediction based on prior knowledge regarding the static and dynamic scenes. The representation is highly interpretable and reusable, but more extension is required to explicitly learn interactive patterns from data.
Research on unsupervised representation learning was also presented recently to learn to identify driving modes in car following scenarios \cite{lin_moha_2018} or extract motion primitives for vehicle encountering \cite{wang_understanding_2018}. The extracted patterns were expected to be more interpretable and reusable, and it is desired to train and test the models via driving data with complete information of surrounding entities in various highly interactive scenarios. In this work, we explicitly provide the answers to \textit{whether}, \textit{when} and \textit{what} for extracting interactive events and patterns with better interpretability and reusability based on highly interactive driving data with complete surrounding information.

\subsubsection{Interactive pattern extraction}
Research efforts have been devoted to modeling interactions of agents. 
\cite{santoro2017simple} constructed a relation network as an augmentation to other modules to handle relational reasoning. 
An attention-based method for multi-agent predictive modeling was presented in \cite{hoshen2017vain}. \cite{shu2018perception} proposed a hierarchical Bayesian model to perceive interactions from movements of shapes. \cite{choi2019learning} constructed a relation-aware framework to infer relational information from the interactions. The aforementioned works are based on either supervised learning of predefined interaction types or implicit modeling of interactions. In this paper, we explicitly extract interactive driving patterns with unsupervised learning to grant insight on how drivers tackle complicated driving scenarios and provide more interpretable representation for downstream modules.

Explicit inference over the potential interactions were performed in \cite{kipf2018neural, webb2019factorised}. Neural Relational Inference (NRI) \cite{kipf2018neural} proposed a novel framework to output explicit interaction types of entities, and \cite{webb2019factorised} further expanded the model to enable the combination of interactions types at the same time step. The interaction type between two objects is static, that is, given trajectories of two objects, NRI only provides one interaction type at the latest time step. The model has to be run over and over again by step to predict the interaction types in a period. Such independence among steps might yield frequent switches of interaction types, which is unrealistic according to \cite{Mu2020IROS}. Additionally, the framework determines the interaction type of a single time step based on the historical trajectories only, but the interaction types can be better inferred via utilizing the entire interaction trajectories (i.e., including both historical and future trajectories, similar as bi-LSTM outperforming LSTM). Our framework dynamically assigns interaction types for each time-step according to an overview of the complete trajectory to overcome the issues.

\subsection{Contribution}
Our contributions can be summarized as follows:
\begin{itemize}
	\item We propose the \textbf{I}nteractive \textbf{D}riving event and pattern \textbf{E}xtraction \textbf{Net}work (IDE-Net) which can extract interactive events and patterns based on trajectory data of naturalistic human driving behavior.
	\item We propose a multi-task learning framework in IDE-Net to leverage the power of connected tasks in learning.
	\item We propose a spatial-temporal block with mixed LSTM and Transformer modules to encode trajectories, which achieves the best performance in experiments.
	\item We find three explainable interaction patterns by IDE-Net, which brings insight for driver behavior modeling. We did both objective and subjective evaluation to analyze the learned patterns. We also show that the proposed model could generalize well across scenarios with different road conditions and coordinate system.
\end{itemize}

\section{Problem Formulation}
\label{sec:prob_formulation}

In this work, we aim to design a learning framework which can automatically extract different interaction patterns in human driving behaviors by observing their joint trajectories. In a two-agent setting, a pair of joint trajectories from the two vehicles defines one sample. Suppose that the joint trajectories are of length $T$, i.e., $T$ time steps. Then a sample contains $(D^1, D^2)$ where $D^1=\{\bm{d}^1_1, \bm{d}^1_2, ..., \bm{d}^1_T\}$ and $D^2=\{\bm{d}^2_1, \bm{d}^2_2, ..., \bm{d}^2_T\}$ are, respectively, the trajectory data of vehicle $1$ and vehicle $2$. We also denote $D_t = \{\bm{d}^1_t, \bm{d}^2_t\}$  as the set of two vehicles' features at time step $t$. With these definitions, our goal is build a learning structure (such as a deep neural network) which takes in $(D^1, D^2)$ and output interaction patterns at each time step, i.e., $l_{\text{type}} = \{l_{1, \text{type}}, l_{2, \text{type}}, ..., l_{T, \text{type}} \}$. No-interaction is also counted as one special interaction type. 

Since there are no golden criteria regarding the interaction patterns, the proposed task is thus an unsupervised learning task without direct-accessible labels. In \cref{sec:model}, we will introduce some unique network structures which helps us achieve such a goal.

\begin{figure*}[!h]
    \centering
    \includegraphics[width = 0.8\textwidth]{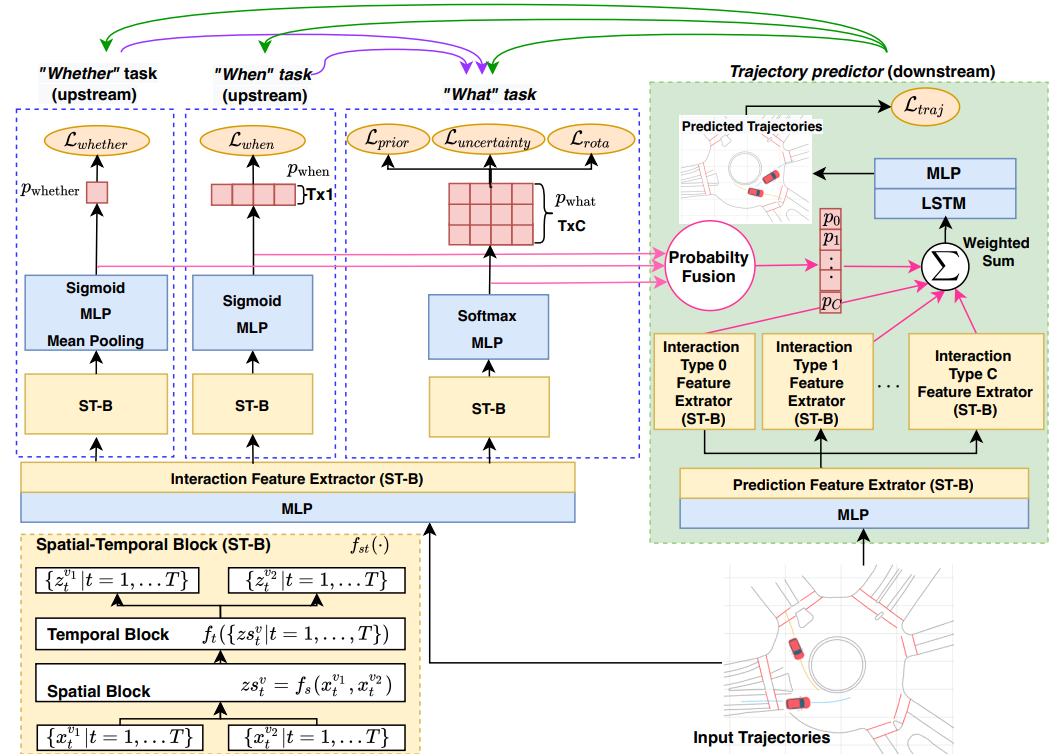}
    \caption{The overview structure of the proposed IDE-Net. Its main task is to extract interactive driving patterns from human data, i.e., the unsupervised "\emph{what}" task. To assist such an unsupervised learning task, we designed the IDE-Net to contain three more auxiliary tasks: a supervised "\emph{whether}" task to predict interaction happens between the two vehicles, a supervised "\emph{when}" task to predict the interacting time period, and finally a self-supervised trajectory prediction (TP) task (green box). The "\emph{whether}" and "\emph{when}" tasks are upstream tasks, thus they can utilize easy-to-obtain labels to extract low-level features that are shared with the "\emph{what}" task. The TP task is a downstream task, which allows the IDE-Net to leverage the influence from the interaction types to the joint trajectories to better extract interaction patterns. For all the four tasks, we use several Spatial-Temporal Blocks (ST-B) $f_{st}(\cdot)$ to encode the spatial-temporal trajectories. MLP means multilayer perceptron. 
   \label{fig:overview}}
\end{figure*}

\section{Model Architecture}\label{sec:model}

\subsection{Trajectory Encoding via Mixed LSTM and Transformer\label{sec:st-blcok}}
Note that interactive driving patterns emerge from the joint trajectories of two interacting vehicles. To encode such spatial-temporal signals, we propose a spatial-temporal block (SB-T), as shown in the yellow box in \cref{fig:overview}. Each ST-B contains one spatial block and one temporal block, sequentially connected. For the spatial block, Transformer layers \cite{vaswani2017attention} are introduced to obtain agent-permutation-invariant representation of the joint trajectories while fusing the relational information of two agents at each time step\footnote[1]{Transformer is an attention-based order-independent set operation in deep learning. It can effectively fuse information in set in a pairwise way and has obtained huge success in the NLP field \cite{devlin2018bert, radford2019language}.}. Such fused representation will be further encoded by the temporal block that contains several LSTM layers \cite{hochreiter1997long}. The relationship between the input and output of the ST-B can be represented by: 
\begin{equation}\label{equ:st-block}
    [Z^{v_1}, Z^{v_2}] =  f_{\text{st}}([X^{v_1}, X^{v_2}]),
\end{equation}
where $[X^{v_1}, X^{v_2}]$ is the input representation of the joint trajectories, and $[Z^{v_1}, Z^{v_2}]$  are correspondingly the output representation. Note that temporally, the lengths of the input and output equal. Therefore, several ST-Bs can be stacked to encode complex dynamics of the joint trajectories.

\subsection{Learning via Auxiliary Tasks}

To tackle the unsupervised "\emph{what}" task, we design three auxiliary self/- supervised tasks in the IDE-Net to help extract interaction patterns. As shown in \cref{fig:overview}, the three tasks include: 1) an upstream "\emph{whether}" task to predict whether interaction occurs between two vehicles, 2) another upstream "\emph{when}" task to predict at which time steps such interaction occurs, and 3) a downstream trajectory prediction (TP) task to predict the joint future trajectories based on historical observations and the predicted interaction probabilities from the "\emph{whether}", "\emph{when}" and "\emph{what}" tasks. We call the "\emph{whether}" and "\emph{when}" tasks as upstream tasks because interaction patterns will exist only when interaction happens, and the features used to encode the upstream tasks should be considered in the pattern extraction task. Therefore, we design a shared "Interaction Feature Extractor" for all the three "\emph{whether}", "\emph{when}" and "\emph{what}" tasks in IDE-Net. Similarly, we call the TP task as a downstream task because different interaction patterns will influence the final trajectories of the vehicles, namely, the joint trajectories are effective indicators of interaction types. Via the "probability fusion" block and the weighted feature extractors in the TP task in the IDE-Net (\cref{fig:overview}), we allow such information flow. Hence, IDE-Net leverages the power of relatively easy but connected upstream and downstream tasks to extract unknown interaction patterns. Both the "\emph{whether}" and "\emph{when}" tasks are supervised tasks where the ground-truths are manually labeled via rules (see Appendix \cref{app:rule} for detailed rules). 
The TP task is self-supervised since all the ground-truth future joint trajectories can be directly observed from data itself.

\subsection{Unsupervised "\emph{What}" Task - Interaction Pattern Extraction}
In the "\emph{What}" task, we assume that there are $C$ types of interaction patterns between two agents, and let network to predict the probability of each type at each time step. Although the three auxiliary tasks help reduce the difficulty of the "\emph{What}" task, we cannot rely on those tasks due to the mode collapse problem. Specifically, the mode collapse problem means that the model would always tend to output only one kind of interaction type because the neural network's easiest strategy to minimize the trajectory prediction loss is to output only one type and optimize its corresponding parameters in the Trajectory Predictor, which is not what we want. Therefore, in IDE-Net, we introduce three additional loss to encourage it to extract meaningful interaction patterns, namely, the prior loss $\mathcal{L}_{\text{prior}}$, uncertainty loss $\mathcal{L}_{\text{uncertainty}}$, and rotation-invariant loss $\mathcal{L}_{\text{rota}}$, as shown in \cref{fig:overview}.

\paragraph{Prior Loss.} To alleviate mode collapse, we make an assumption that different interaction types approximately distribute uniformly in the dataset. Therefore, we would punish prediction outputs that are significantly deviating from such priors, i.e., a prior loss $\mathcal{L_{\text{prior}}}$ is given as the entropy of the interaction type distribution:
\begin{equation}\label{equ:prior-loss}
    \mathcal{L_{\text{prior}}} = \sum\limits^C p_{c}\log p_{c}, \text{  with  } p_c = \frac1{NT}\sum\limits^N\sum\limits^T p^t_{\text{what},n,c}.
\end{equation}

\paragraph{Uncertainty Loss.}
If only with prior loss, the model finds another way to cheat: it tends to output equal confidences for all kinds of interactions at all time-steps of all samples, which can minimize the prior loss but is not what we want as well. Therefore, to encourage the IDE-Net to output certain prediction at each time step for some interaction type $c$, an uncertainty loss is introduced to punish outputs with 
high entropy at each time step:
\begin{equation}\label{equ:uncertain-loss}
    \mathcal{L_{\text{uncertainty}}} = \frac1{NT}\sum\limits^N\sum\limits^T\sum\limits^C -p^t_{\text{what},n,c}\log p^t_{\text{what},n,c}.
\end{equation}
By using the uncertainty term $\mathcal{L_{\text{uncertainty}}}$,  the model would tend to output $\{[1,0,0], [0,1,0], [0,0,1]\}$ instead of simply $\{[0.333,0.333,0.333], [0.333,0.333,0.333], [0.333,0.333,0.333]\}$.

\paragraph{Rotation-Invariant Loss.}
Inspired by recent success of contrasting learning \cite{chen2020improved,chen2020big}, we would like to make the prediction of interaction types rotation-invariant. For each sample, we randomly rotate it twice with different angles and punish inconsistent outputs:
\begin{equation} \label{equ:rotation-inv loss}
    \mathcal{L_{\text{rota-Invariant}}} =\frac1{NTC}\sum\limits^N\sum\limits^T\sum\limits^C (p^{t,1}_{\text{what},n,c} - p^{t,2}_{\text{what},n,c})^2,
\end{equation} where $p^{t,1}_{\text{what},n,c}$ and  $p^{t,2}_{\text{what},n,c}$ represent the confidences of interaction type $c$ at time-step $t$ predicted based on the same sample $n$ with two rotation angles respectively.

\begin{table*}[!h]
    \scriptsize
	\begin{tabular}{c|c|cc|cc|cc|cc|cc}
		\hline
		\multirow{2}{*}{Method}      &   \multirow{2}{*}{Setting}     & \multicolumn{2}{c|}{FT}                     & \multicolumn{2}{c|}{GL}                     & \multicolumn{2}{c|}{MA}                     & \multicolumn{2}{c|}{SR}                     & \multicolumn{2}{c}{EP}                     \\  \cline{3-12}
		&          & When                 & Whether              & When                 & Whether              & When                 & Whether              & When                 & Whether              & When                 & Whether              \\ \hline
		\multirow{3}{*}{LSTM}        & Single   & 0.818                & 0.926                & 0.914                & 0.923                & 0.783                & 0.862                & 0.500                & 0.891                & 0.904                & 0.889                \\  
		& Transfer & 0.491                & 0.798                & 0.599                & 0.724                & 0.635                & 0.793                & 0.615                & 0.864                & 0.735                & 0.865                \\  
		& Finetune & {\ul 0.837}          & {\ul \textbf{0.912}} & {\ul 0.937}          & {\ul 0.924}          & {\ul 0.815}          & {\ul \textbf{0.871}} & {\ul 0.750}          & {\ul 0.923}          & {\ul 0.923}          & {\ul \textbf{0.944}} \\ \hline
		\multirow{3}{*}{Transformer} & Single   & 0.740                & 0.877                & 0.862                & 0.884                & 0.727                & 0.839                & 0.147                & 0.897                & 0.333                & 0.846                \\  
		& Transfer & 0.433                & 0.777                & 0.528                & 0.718                & 0.535                & 0.760                & 0.473                & 0.884                & 0.612                & 0.841                \\  
		& Finetune & {\ul 0.751}          & {\ul 0.882}          & {\ul 0.883}          & {\ul 0.901}          & {\ul 0.741}          & {\ul 0.876}          & {\ul 0.559}          & {\ul 0.910}          & {\ul 0.778}          & {\ul 0.923}          \\ \hline
		\multirow{3}{*}{Mixed}       & Single   & 0.848                & {\ul 0.911}          & 0.918                & 0.922                & 0.839                & 0.859                & 0.706                & 0.936                & 0.907                & 0.892                \\  
		& Transfer & 0.542                & 0.812                & 0.681                & 0.692                & 0.634                & 0.802                & 0.637                & 0.879                & 0.653                & 0.833                \\  
		& Finetune & {\ul \textbf{0.878}} & 0.909                & {\ul \textbf{0.946}} & {\ul \textbf{0.929}} & {\ul \textbf{0.853}} & {\ul 0.869}          & {\ul \textbf{0.882}} & {\ul \textbf{0.949}} & {\ul \textbf{0.930}} & {\ul \textbf{0.944}} \\ \hline
	\end{tabular}
    \caption{Prediction performance comparison of three ST-B settings.  FT, GL, MA, SR and EP are different scenarios in the INTERACTION dataset.\vspace{-3mm}}
    \label{tab:st-block}
\end{table*}

\section{Experiments}
\subsection{Datasets and experiment settings}
Trajectory datasets facilitating meaningful interaction extraction research should contain the following aspects: 1) a variety of driving scenarios with complex driving behaviors, such as roundabouts, unsignalized intersections, merging and lane change, etc.; 2) densely interactive driving behavior with considerable numbers of interaction pairs; 3) HD map with semantic information; 4) complete information (without occlusions) of the surrounding entities which may impact the behavior of the investigated objects. The INTERACTION dataset \cite{interactiondataset} meets all the aforementioned requirements, and our experiments were performed on it. We randomly selected $15657$ pairs of vehicles from five different driving scenarios as the training set, and another $1740$ pairs as the test set. We empirically set the number of interaction types as three\footnote{We performed user studies on the results of different number of the interaction types. Three is the number with the best performance.}, and it serves as a hyper-parameter of our model.

\subsection{Data Preprocessing}
Note that in structured driving environments, vehicle trajectories differ significantly among different HD maps. Our goal is to extract the interactive driving events and patterns between agents in a generic fashion over all such scenarios, i.e., excluding the influences of different road structures. Therefore, we proposed three data preprocessing methods to reduce such influences from road structures:

\begin{itemize}
    \item  Coordinate Invariance (CI): CI converts global coordinates to relative coordinates by setting the origin of coordinate system as the center position of the two agents' initial positions.
    
    \item Orientation Invariance (OI): OI helps to make features not sensitive to absolute orientations by data augmentation with random rotation at each batch.
    
    \item Rotation Normalization (RN): RN normalizes the coordinates in different scenarios according to scale of the augmented data in OI. To make the rotation valid and keep the covariance of the input feature to be $1$, it applies the same scale factor $\sqrt{E\frac{(X)^2+(Y)^2}{2}}$ to both $X$ and $Y$ coordinate.
    Details regarding the RN step is given in the Appendix \cref{app:RN}. 
\end{itemize}

Appendix \ref{app:ablation-processing} gives the ablation study for the three methods.

\subsection{Performance of the "When" and "Whether" Tasks}
\paragraph{Manipulated Factors.} We manipulated a single factor: the structure of the ST-B in IDE-Net. Three conditions were compared: 1) LSTM layers only, 2) Transformer layers only, and 3) the proposed mixed LSTM and Transformer in \cref{sec:st-blcok}. In setting 2), we add the position embedding similar to \cite{vaswani2017attention} to capture the temporal signals.

\paragraph{Dependent Variables.} For the "\emph{whether}" task, we measured the prediction accuracy. For the "\emph{when}" task, we measured the intersection-over-union (IoU) accuracy and adopt \textbf{IoU 0.6 accuracy} as correct prediction, namely, if IoU of a prediction is larger than 0.6, it counts as an accurate prediction. IoU is defined as:
\begin{equation}\label{equ:IoU}
    \text{IoU} = \frac{[\hat{l}_{\text{start}},\hat{l}_{\text{end}}] \cap [l_{\text{start}}, l_{\text{end}}]}{[\hat{l}_{\text{start}},\hat{l}_{\text{end}}] \cup [l_{\text{start}}, l_{\text{end}}]},
\end{equation}
where $\hat{l}_{\text{start}}$ and $\hat{l}_{\text{end}}$ are, respectively, the predicted starting and ending frames of interactions.

\paragraph{Hypothesis.} We hypothesize that the proposed mixed LSTM and Transformer ST-B structure can achieve better prediction accuracy in "\emph{whether}" and "\emph{when}" tasks compared to the other two settings.

\paragraph{Results.} We tested the prediction performances on five different scenarios (FT, GL, MA, SR, EP) in the INTERACTION dataset \cite{interactiondataset} with diverse number of samples ranging from $9000$ to $100$. We tested the performance under three different training settings: 1) \emph{Single} means training and testing both with the current one, 2) \emph{Transfer} means training with all the other scenarios and testing on the current one, and 3) \emph{Finetune} means training with all the other scenarios and then finetuning with the current one. 

The results are shown in \cref{tab:st-block}. We can conclude that: 1) "\emph{whether}" task was much simpler than "\emph{when}" task since even training without samples of the current scenario in the dataset, it was still able to achieve decent results ($>0.7$); 2) regarding the prediction performance, Mixed $>$ Pure LSTM $>$ Pure Transformer, which shows the effectiveness of Transformer for set of data and LSTM for sequence of data; and 3) finetuning always had the best performance among the three settings. It proved that better generalization ability can be achieved by taking into account more samples from other driving scenarios which may have completely different road structures. Therefore, for all the following experiments, we used the proposed mixed ST-B as it achieved the best performance among the three compared structures.

\begin{figure*}[!h]
    \subfloat[Noticing (Purple)-Yielding (Green)-Caution (Blue). For simplicity, we denote one vehicle as $v1$ and the other as $v2$. The interaction began when the two vehicles noticed each other (Purple). Then $v2$ yielded and $v1$ kept going (Green). Finally, $v2$ started to proceed slowly after $v1$ passed and the interaction ended.]{\includegraphics[width = 0.99\linewidth]{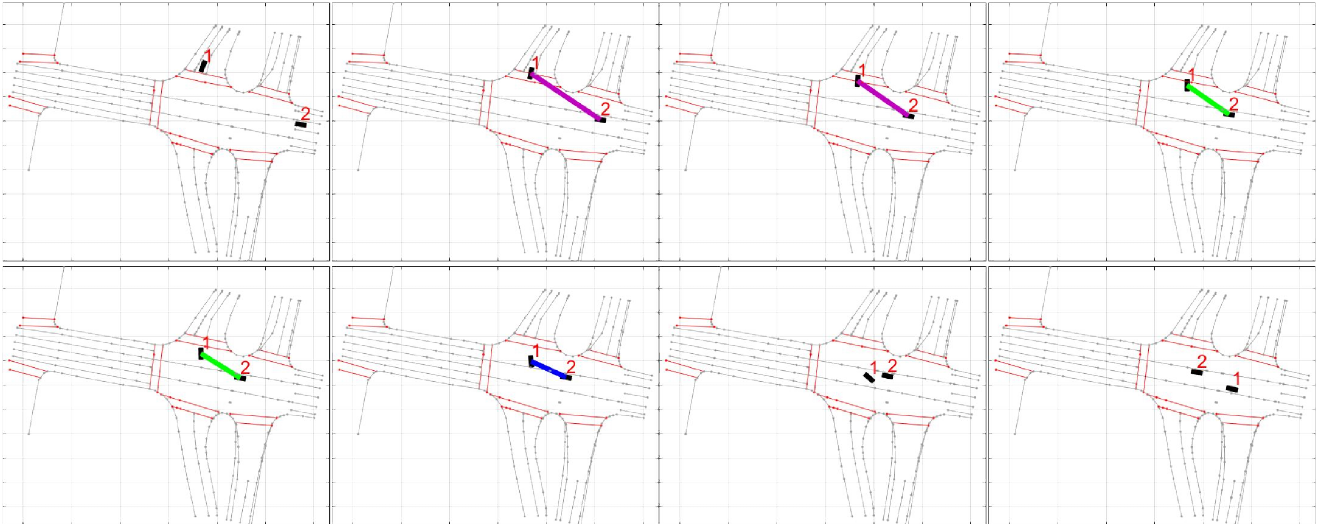}} 
    \par\medskip
    \subfloat[Yielding (Green)-Caution (Blue). Before $v1$ arrived, $v2$ was waiting at the stop sign. There was no Noticing part and $v1$ just passed while $v2$ kept yielding (Green). Once $v1$ passed $v2$, $v2$ proceeded immediately and followed $v1$ closely (Blue). Finally, the interaction ended.]{\includegraphics[width = 0.99\linewidth]{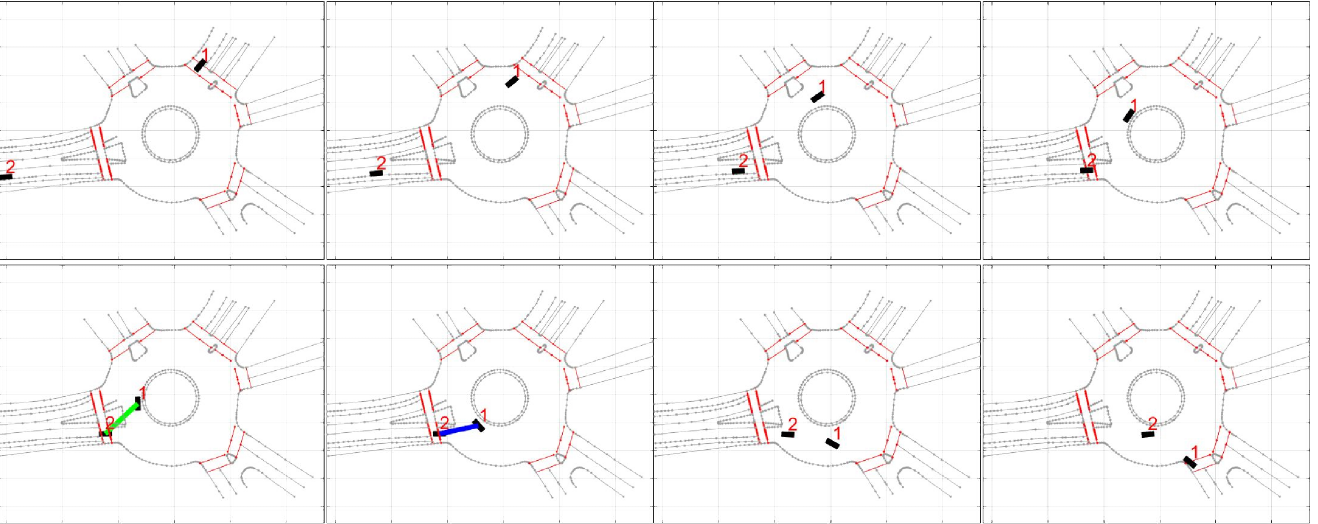}}
    \caption{Examples of interaction patterns.}
\label{fig:interaction-vis1}
\end{figure*}

\begin{figure*}[!h]
    \subfloat[Yielding (Green). $V1$ stopped before $v2$ arrived and thus there was no noticing part. $V1$ proceeded when $v2$ passed for a while. There was no potential collision (no caution part). This example shows that the model was able to capture patterns of vehicle motions separately.]{\includegraphics[width = 0.99\linewidth]{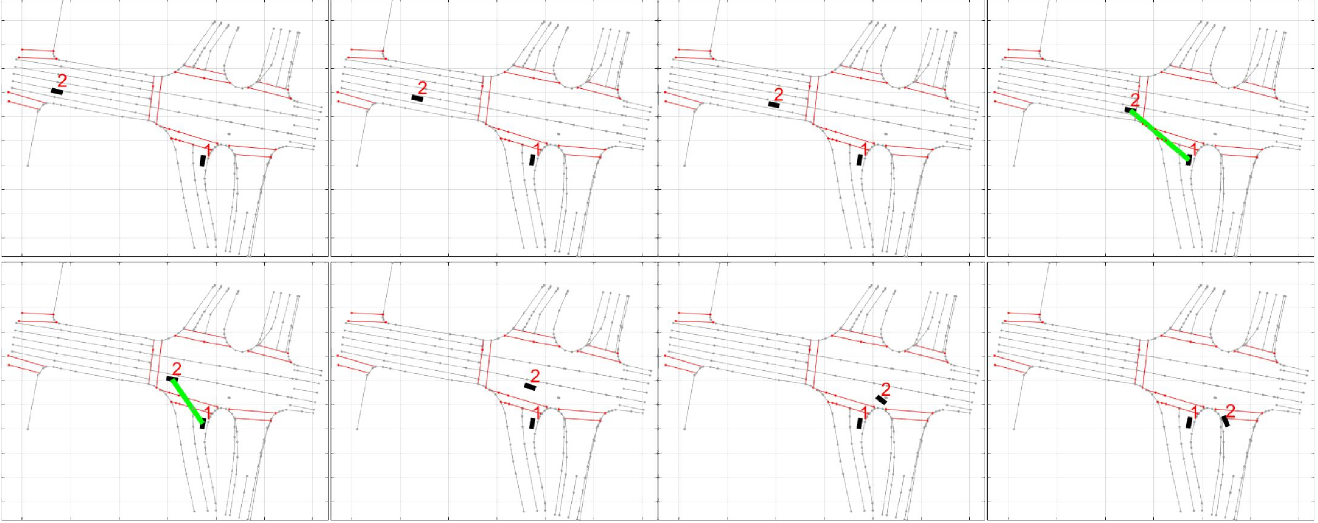}}
    \par\medskip
    \subfloat[Noticing (Purple)-Yielding (Green). Both vehicles noticed each other at a very early stage and slowed down. $V2$ yielded until $v1$ passed. There was caution part because their directions were perpendicular and once one of the vehicles passed, it was impossible to collide and thus there was no need to be overly cautious.]{\includegraphics[width = 0.99\linewidth]{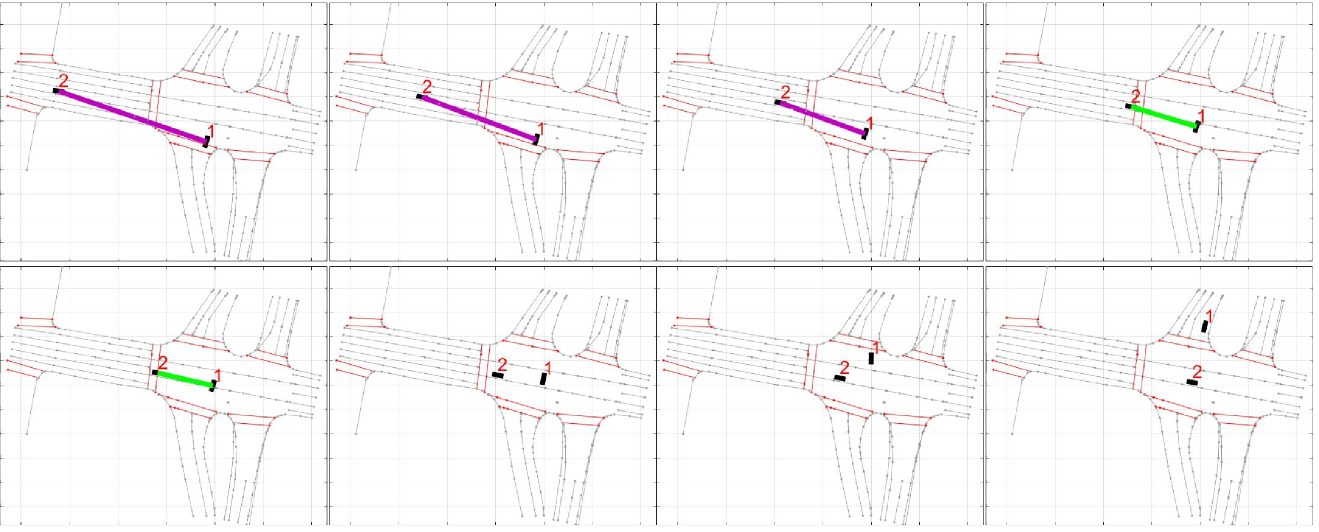}}
    \caption{Examples of interaction patterns.}
\label{fig:interaction-vis2}
\end{figure*}

\subsection{Results of the "\emph{What}" Task}

\subsubsection{Learned Patterns}

From the interaction types given by the IDE-Net, we conclude the three following patterns (interaction type):
\begin{itemize}
    \item \textbf{Noticing}: two vehicles observe each other and realize that there would be overlaps between their paths. They slow down to avoid collisions, although the speeds may still be relatively high.
    \item \textbf{Yielding}: one vehicle tends or decides to yield. 
    \item \textbf{Caution}: two vehicles are relatively close to each other; one moving and the other may also be moving but cautiously.
\end{itemize}

Figures \ref{fig:interaction-vis1} and \ref{fig:interaction-vis2} demonstrate cases of the learned interaction types at different time steps of interactive trajectory pairs.  We uniformly sampled $8$ frames to represent each example video. In each subfigure, the line between the two vehicles represents that they are interacting and different colors represent different interaction types (Noticing-Purple, Yielding-Green, Caution-Blue).

\subsubsection{Quantitative Evaluation}

To quantify the performance of the pattern extraction task (\emph{what} task), we adopted two metrics: 1) a cross-validation metric which compares the distribution difference between the extracted pattern types in the training and test sets; and 2) a user study which compares the human labeled interaction patterns with the extracted ones via the IDE-Net.

\begin{figure*}[!h]
    \centering
    \includegraphics[width = \textwidth]{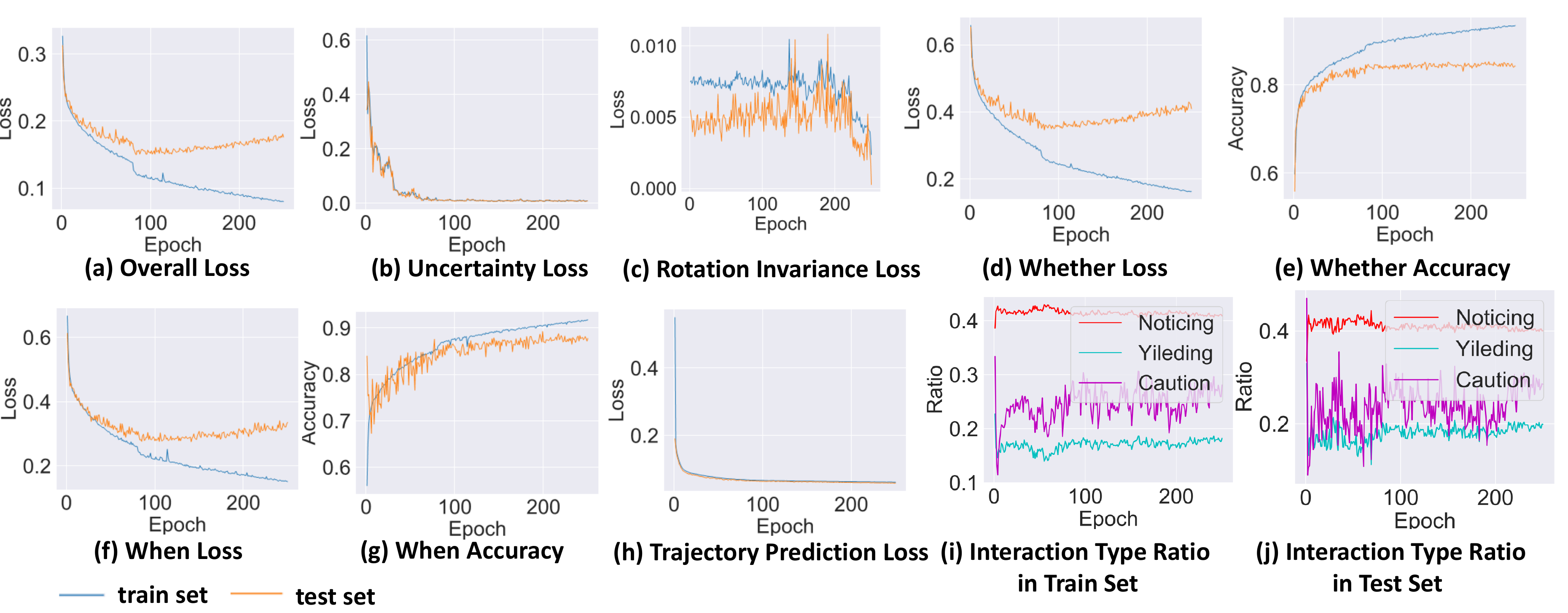}
    \caption{An overview of the performance of IDE-Net}
\label{fig:loss-change}
\end{figure*}

\begin{table}[!ht]
\caption{Results with both the cross-validation metric and the user study\label{tab:2}}
\subfloat[Ratios of each interaction type on training and testing sets\label{tab:type-ratio}]{\begin{tabular}{c|c|c|c}
        \hline
        Type & Noticing & Yielding & Caution \\ \hline
        Train       & 0.469  & 0.208  & 0.322  \\
        Test   & 0.452  & 0.229  & 0.322  \\ \hline
        \end{tabular}} \quad\quad\quad\quad
\subfloat[Results of user study 
\label{tab:user-study}]{\begin{tabular}{c|c|c|c}
        \hline
        VTA & MTA & VAA &  MAA \\ \hline
         92.5\%            & 76.0\% &   100.0\%  &90.5\%   \\ \hline
        \end{tabular}}
\end{table}

\paragraph{Results with the Cross-Validation Metric.}
The percentages of the extracted interaction patterns are shown in \cref{fig:loss-change} (i)-(j) and \cref{tab:2} (a). The distribution of the three interaction patterns on the training and test sets were quite similar, which indicates good generalization capability of the proposed IDE-Net, i.e., it did not overfit the training set but indeed learned to extract different interaction patterns of vehicles.  

\paragraph{User Study}
We also conducted an user study to provide a subjective metric to evaluate the performance of IDE-Net. We recruited $10$ users with driving experiences, $5$ males and $5$ females. We visualized the interactive trajectories as well as the extracted interaction types together as videos and showed each user $12$ videos (each scenario has at least $2$ videos). We asked the users to label 1) one of the three interaction types by themselves, and 2) whether they agree with the predicted interaction types.
The procedure of the user study is as follows:
\begin{enumerate}
    \item Initial setting: We told users the meaning of the three interaction types: Noticing (N), Yielding(Y), and Caution(C). Each user watched all $12$ videos in a random order.
    \item Experiment I: We showed users videos only with marks indicating whether the two vehicles were interacting at each time step. We asked users to provide the interaction types and their order.
    \item Experiment II: We showed them videos with marks indicating the interaction types at each time step. We asked users whether they agree with the types labeled by IDE-Net.
\end{enumerate}

\textbf{Metric}: In Experiment I, we considered the interactions types labeled by the model were aligned with user inputs if both the interaction types and their order were the same. We calculate two metrics based on it: 1) for each video, we averaged the human labels by voting (Voting Type Accuracy - VTA); and 2) we took the mean accuracy over all videos and all users (Mean Type Accuracy - MTA). In Experiment II, similarly, we defined two metrics: \textit{Voting Agreed Accuracy - VAA} and \textit{Mean Agreed Accuracy - MAA}, which were obtained via the similar process as the \textit{Voting Type Accuracy} and \textit{Mean Type Accuracy} in Experiment I.

The results are shown in \cref{tab:2}(b). The interaction types labeled by the users aligned quite well with the results obtained by the IDE-Net. This shows that IDE-Net was able to extract the generic interaction patterns and the extracted patterns are quite interpretable.

\subsection{Analysis and Discussions}

\begin{figure}
    \centering
    \includegraphics[width = 0.7\textwidth]{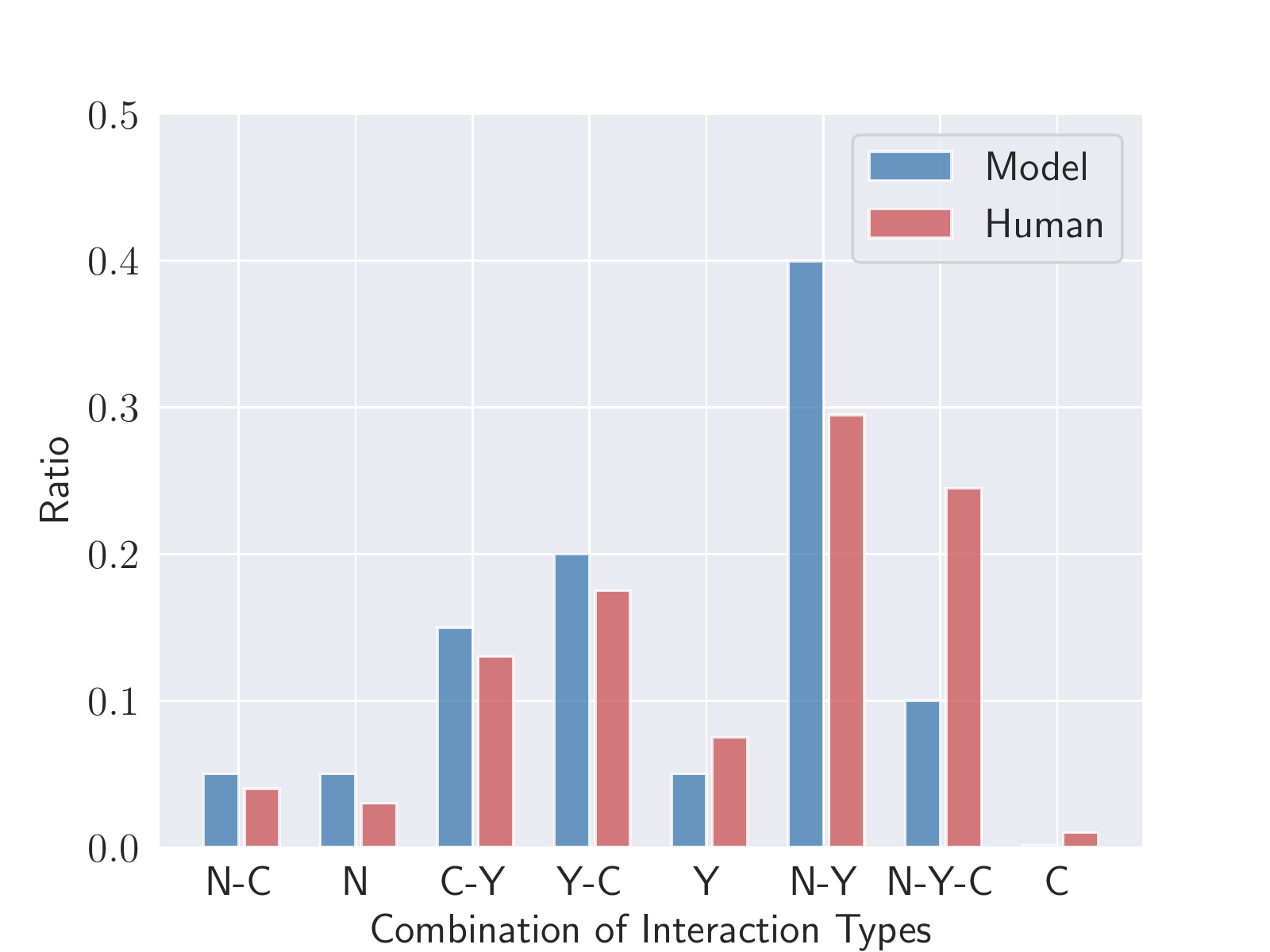}
    \caption{The ratio of permutations labeled by the model and by human. \vspace{-6mm}}
     \label{fig:ratio-permutation}
\end{figure}

\begin{table}[]
    \centering
	\begin{tabular}{c|c|c|c}
		\hline
		& Noticing & Yielding & Caution \\ \hline
		Model     & 0.60          & 0.90 &   0.50 \\
		Human & 0.61            & 0.92 &    0.60    \\ \hline
	\end{tabular}
    \caption{\vspace{3mm}The frequency of interaction types.\vspace{-6mm}}
    \label{tab:avg-video}
\end{table}

The performance of IDE-Net over all losses is summarized in \cref{fig:loss-change}(a)-(h). We can see that although the supervised "\textit{whether}" and "\textit{when}" tasks show a sign of over-fitting, the uncertain loss kept decreasing and no over-fitting occurred for the trajectory prediction task. Moreover, the consistent decreasing in uncertain loss means that the model became more and more certain about the interaction types of each time step. We also counted the time steps when the confidence over a specific interaction type was significantly larger than the other two, i.e., the confidence was greater than $0.9$ and divided it by the overall length of the interaction window. We find that the ratio was more than \textbf{99.5\%}, which indicates that the proposed model was able to find the significant differences among interaction types.

We also calculated the frequency of different interaction types. As shown in \cref{tab:avg-video}, the results given by IDE-Net were quite close to that given by human, although human tended to label more interaction types. It means that IDE-Net was relatively conservative.

To explore the co-appearance relationship of different interaction types, we define the interaction types and the order of their appearance as a permutation. For instance, in a video, if the two vehicles are labeled as first Noticing, then Yielding, and finally Caution, the permutation of this prediction is then "N-Y-C". We calculated the ratio of permutations labeled by the model and by users as shown in \cref{fig:ratio-permutation}. Human behaved quite similar to the model except that human tend to give more "N-Y-C". "N-Y-C" was preferred probably due to the biased default in human, i.e., all three interaction types should appear and show up in sequence.

\subsection{Differences among Learned Interaction Patterns}

\begin{figure}[!h]
    \vspace{-6mm}
    \centering
	\subfloat{\includegraphics[width = 0.5\linewidth]{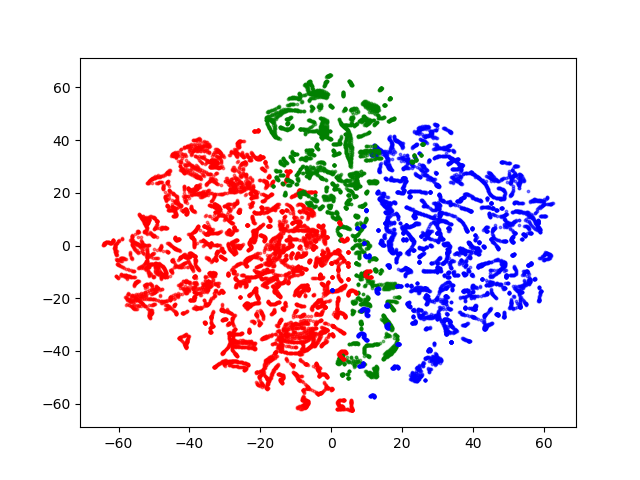}} 
	\subfloat{\includegraphics[width = 0.5\linewidth]{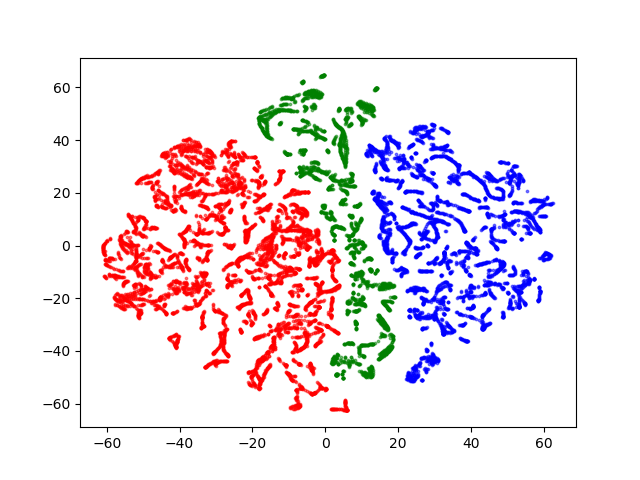}}
	\caption{Visualization for features of time-steps by t-SNE. Each color represents a type of interaction. Features before fed into the output MLP for the \emph{What} task (subfigure left). Features before fed into the LSTM for the Trajectory Prediction task (subfigure right).}
	\label{fig:feature-visualization}
\end{figure}

\begin{figure}[!h]
    \vspace{-6mm}
    \centering
	\includegraphics[width = \textwidth]{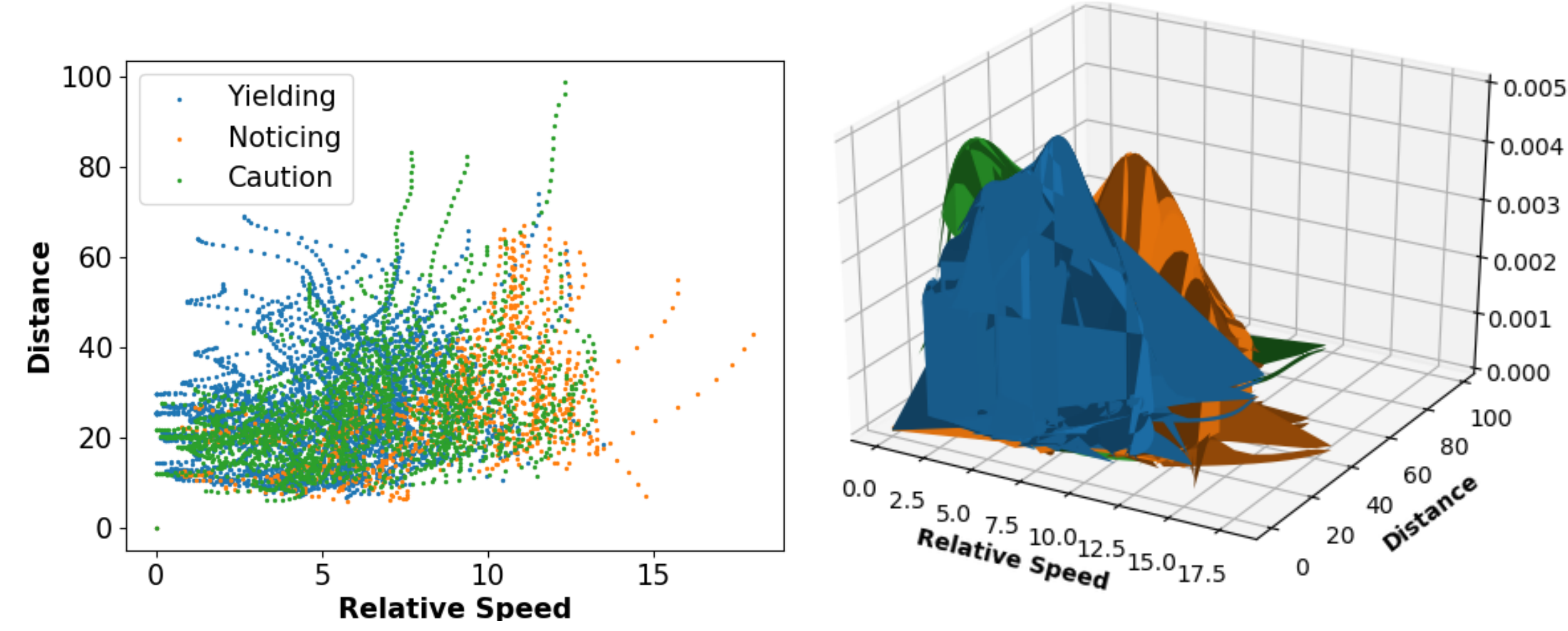}
	\caption{Orange-Noticing, Blue-Yielding, Green-Caution. Visualization on relative distance - relative speed space with each color representing a interaction type (subfigure left).  Its probability density distribution estimated by Gaussian kernel density estimation (subfigure right).}
	\label{fig:feature-visualization2}
\end{figure}

There may be concerns that the interaction types and their corresponding features for trajectory prediction were very similar to each other. Thus, in this section, we checked the differences among the three learned interaction patterns. We visualized the the hidden layer before the output MLP layer for the \emph{What} task, as well as the hidden layer before the LSTM module for the Trajectory Prediction task. As shown in Fig.~\ref{fig:feature-visualization}, clusters of interaction patterns are linearly separable in both figures. The results show that the model can classify interaction types with high confidences and predict trajectory in a interaction-type-specific way.

Additionally, we visualize the interaction types of each time-step in a feature space including the relative distance and relative speed. In Fig.~\ref{fig:feature-visualization2} (left), their distributions were different. The characteristics of three interaction types are as follows: "Noticing" (Orange) happens when the two vehicles are driving towards a potentially conflicting area but still at relatively larger distances compared to "Yielding" (Blue) and relatively larger speeds compared to "Caution" (Green); "Yielding" features smaller relative distances; and "Caution" features the smallest relative speed, indicating negotiation.  

To have a better illustration of the differences of their distributions, we conducted the kernel density estimation to estimate the probability density function 
As shown in Fig.~\ref{fig:feature-visualization2} (right), the three interaction types has different peaks, which shows that our model can capture the difference of different interactions patterns considering both relative speed and distance.

\section{Ablation Study}

IDE-Net aims to extract interpretable interaction types between vehicles. Towards the goal, the proposed "\textit{whether-when-what}" paradigm has two benefits. First, the three tasks are intrinsically hierarchical with semantic meanings and increasing complexity. The existence of upstream tasks can improve the performance of downstream tasks. Second, the feature-sharing structure leverages the power of multi-task learning to boost both the learning performance and efficiency. We conducted ablation studies to support these two statements. 

We conducted several comparison experiments: 1) only \textit{whether}  task, 2) only \textit{when}  task, 3)  only \textit{whether} and \textit{when}  tasks, and 4) our approach. Under each experiment setting, we calculate the accuracy (Acc) of the \textit{whether} and \textit{when} tasks. The goal of this set of comparison experiment is to demonstrate the power of feature-sharing structure in improving the learning efficiency and performance.

Results are shown in Tab. \ref{tab:ablation-whether-when}. Under the multi-task learning framework, performances of both \emph{whether} and \emph{when} tasks were improved remarkably. Note that the Whether Acc was bad with \emph{when} task only. The reason is that compared to the binary classification in "whether" task, it is much harder for \emph{when} task to output all $0$s for the non-interactive trajectories due to the probabilistic nature of deep learning. Additionally, the parallel training of multiple tasks did not make a significant difference in terms of converging time (best epoch) compared to training those tasks separately. Therefore, the multi-task learning framework, which was able to avoid the burden of learning different low-layer features for each task, is efficient.

\begin{table}[]
\footnotesize
\begin{tabular}{cccccc}
\hline
\textbf{Whether}          & \textbf{When}             & \textbf{What}             & \textbf{Whether Acc} & \textbf{When Acc} & \textbf{Best Epoch} \\ \hline
\checkmark &                           &                           & 0.893                  & -   & 264                   \\ 
                          & \checkmark &                           & 0.181                  & 0.843  & 279                \\ 
\checkmark & \checkmark &                           & 0.898                  & 0.867      & 263            \\ 
\checkmark & \checkmark & \checkmark & 0.909                  & 0.878    &     291          \\ \hline
\end{tabular}
\caption{Ablation study for \emph{whether} and \emph{when} tasks under different multi-task settings. Note that with \emph{when} task only, if the output sequence contains no "$1$" labels, the output of \emph{whether} task is assigned as "$0$".\vspace{-6mm}}
\label{tab:ablation-whether-when}

\end{table}

We have also conducted another user study for the results with \emph{what} task only. The users cannot distinguish the extracted interaction patterns. This proved that the proposed multi-task structure can improve the interpretability of the learned patterns.

\section{Conclusion}
In this paper, we proposed a framework to automatically extract interactive driving events and patterns from driving data without manual labelling. Experiments on the INTERACTION dataset across different driving scenarios were performed. The results showed that through the introduction of the Spatial-Temporal Block, the mutual attention and influence between trajectories of agents were effectively extracted, as evaluated via both objective and subjective metrics. Moreover, we found that the extracted three interaction types are quite interpretable with significantly different motion patterns. Such a framework can greatly boost research related to interactive behaviors in autonomous driving, e.g., interactive behavior analysis and interactive prediction/planning, since it can be used to as a tool to extract not only the interactive pairs but also the interaction patterns from enormous of detected trajectories.

There are still limitations about this work. Currently, the proposed framework focuses on two-agent settings, and for multi-agent settings, we have to enumerate all possible combinations. In real world, however, interactions may not be just between two vehicles, but can be potentially among multi-agents (more than two), and we will extend towards that direction in our future work.



\appendices

\section{Comparison with Neural Relational Inference}

Suppose there are $T$ time-steps in the joint trajectories and the Trajectory Prediction Task is to predict $k$ time-steps. Under NRI's setting for each sample, they only do trajectory prediction for the time-step $T-k$. While in IDE-Net, trajectories are predicted for each time-step $t$ ($t\leq T-k$). The advantage of latter way is that there may be different interaction types at different time steps, which may have different impact on the future motion of vehicles. By predicting at each time step, the impact of interaction type can be fully considered. In contrast, when predicting in the former way, what happened before $T-k$ may not have much influence on their future motions. As a result, the impact of interaction may be degraded or even lost.

Additionally, to avoid data leakage in the downstream trajectory prediction task, we employ causal mask (mask for future time steps) in the trajectory predictor. However, in the upstream tasks and \emph{what} task, all blocks do not have causal mask, which means that each time step can access the features of all the other time steps. There are two reasons not to have causal mask. First, interactions are over a period of time and it is hard to decide the interaction type of a single time step without the entire trajectory. Second, with the proposed structure, only interaction types of each time step are transmitted to the trajectory predictor. If this leakage facilitates trajectory prediction, the predicted interaction types can capture their impact on vehicle motions, which is the goal of our model.

\section{Labeling Rules for Supervised Whether and When Task\label{app:rule}}
In the experiments, we use the INTERACTION Dataset. There are no labels in the original dataset. Thus, for the \emph{whether} task, we assign labels $l_{\text{whether}}$ where 0 represents no-interaction, 1 represents having interaction, and -100 represents not-sure. For the \emph{when} task, we assign labels $l_{\text{start}}$ and $l_{\text{end}}$ to represent, respectively, the start time index and end time index of the interaction period. All the time steps between $l_{\text{start}}$ and $l_{\text{end}}$ are labeled as 1 and otherwise 0. We design two rules to label the samples:
\begin{enumerate}
	\item For each vehicle which stops at the stop sign, if it stops for more than 3 seconds (a hyper-parameter), then for all the vehicles passing through its front zone while it is stopping, we label the pair of vehicles as an interaction pair, i.e., $l_{\text{whether}}=1$. The intuition here is: if a vehicle stops longer than what is required by traffic law, it is highly likely that it is waiting for other vehicles, which should count as an interaction process.
	
	As for $(l_{\text{start}}, l_{\text{end}})$, we label the first time step when the passing vehicle is less than 20 meters (another hyper-parameter) away from the stopping vehicle`s front zone as $l_{\text{start}}$ and the time step that moving vehicle passes the front zone as $l_{\text{end}}$. The reason is that when vehicles are close, there might be highly interactive behaviors to avoid collisions while keeping efficiency.
	
	To give negative samples, when a stopping vehicle stops for less than 1 second (the third hyper-parameter), we label the pair of vehicles as $l_{\text{whether}}=0$ because the stopping vehicle stops so shortly that it might not wait for other vehicles, i.e., no interaction exists. 
	
	For those samples with a stopping vehicle stopping between 1-3 seconds, we label them as not-sure samples. They are not used to calculate the supervised \emph{whether} and \emph{when} loss.
	
	\item For each pair of vehicles, if their minimal time-to-collision (TTC)\footnote[1]{The TTC is defined as the estimated time for the vehicle to arrive at the collision point assuming that the vehicle is running at the constant speed.} interval is less than 3 seconds, we label them as $l_{\text{whether}}=1$. 
	
	We also set the first time step when both vehicles are less than 20 meters (the fourth hyper-parameter) away from their potential collision points as $l_{\text{start}}$ and the first time step when one of them passes the potential collision points as $l_{\text{end}}$. The intuition of this rule is similar: when two vehicles are close to each other, there might be interactions.
	
	Also, we set negative samples as the pairs of vehicles whose minimal time-to-collision time is larger than 8 seconds and not-sure samples as between 3-8 seconds.
\end{enumerate}

Additionally, for pairs of vehicles which have no time-to-collision and none of which stops at a stop sign, they are labelled as $l_{\text{intera}}=0$. For each scenario, we randomly sample those negative samples until the number of positive samples and negative samples are approximately equal.

\section{Statistics of the Dataset}
\begin{table*}[!h]
	\centering
	\footnotesize
	\begin{tabular}{c|c|c|c|c|c|c}
		\hline
		Scenario               & \# Sample & \# Positive Sample & \# Negative Sample & \# Not Sure Sample \\ \hline
		USA\_Roundabout\_SR    & 481       & 182                & 207                & 92                \\ \hline
		USA\_Roundabout\_FT    & 9445      & 3965               & 3978               & 1402              \\ \hline
		USA\_Roundabout\_EP    & 189       & 62                 & 90                 & 27                \\ \hline
		USA\_Intersection\_MA  & 1702      & 734                & 752                & 216               \\ \hline
		USA\_Intersection\_GL  & 5580      & 2281               & 2355               & 944               \\ \hline
	\end{tabular}
	\caption{Statistics of obtained samples.\label{tab:stat-dataset1}.}
\end{table*}

We have 17397 samples from 5 scenarios of the INTERACTION dataset. The statistics is given in Tab.~\cref{tab:stat-dataset1}. Note that we choose the five scenarios since 1. the number of samples is diverse so that we can evaluate the generalization ability and transferring ability of the model. 2. they have different road condition - FT, SR, EP are Roundabouts while GL and MA are intersections.

\section{Details about Rotation Normalization\label{app:RN}}

Since coordinates in different scenarios may have different scales of coordinate, we need to do normalization properly to make the performance of the model generalize well across scenarios. Before that, we would like to first find out the expectation and variance of coordinates after the random rotation in the Orientation Invariance (OI) step. Without loss of generality, we take the $x$ coordinate as an example and it is easy to extend to $y$.  After randomly rotation, the rotated coordinate $x^{\prime} = x * \cos\theta - y * \sin \theta$ where $\theta \sim U(0,\pi)$ is randomly sampled rotation angle for the coordinate system. The goal of RN is to let $E x^{\prime}=0$ and $Dx^{\prime}=1$ where $E(\cdot)$ and $D(\cdot)$, respectively, represent the expectation and variance among all samples and all time steps. Since $x$, $y$, and $\theta$ are independent and $\theta \sim U(0,2\pi)$, we have:
\begin{equation}
\begin{split}
       Ex^{\prime} &= E(x * \cos\theta - y * \sin \theta) \\
       &= \frac{1}{2\pi} E\int_0^{2\pi}x * \cos\theta - y * \sin \theta \mathrm{d}\theta  \\
       &= 0 \\
    \end{split}
\end{equation} which means we do not need to do transnational movement of the coordinate axis in the normalization step. Its expectation is already 0 after random rotation. Then, for variance, we have:
\begin{equation}
   \begin{split}
       Dx^{\prime}_t &= E(x^{\prime})^2 - (Ex^{\prime})^2\\
       & = E(x * \cos\theta - y * \sin \theta)^2\\
       & = \frac{1}{2\pi} E\int_0^{2\pi}(x * \cos\theta - y * \sin \theta)^2 \mathrm{d}\theta\\
       & = E\frac{x^2+y^2}{2}  \\
    \end{split}
\end{equation}
To let $ Dx^{\prime}_t=1$, we divide raw coordinate $x$ by $\sqrt{E\frac{(x)^2+(y)^2}{2}}$. Similarly, since variance equation is symmetric for x and y, we could divide raw coordinate $y$ by the same value. 

\section{Ablation Study for the Three Proposed Data Preprocessing Methods\label{app:ablation-processing}}
\begin{table*}[!h]
	\centering
	\scriptsize
	\begin{tabular}{c|c|cc|cc|cc|cc}
		\hline
		\multirow{2}{*}{Method}      &      \multirow{2}{*}{Setting}    & \multicolumn{2}{c|}{Zero Norm} & \multicolumn{2}{c|}{No Rotation} & \multicolumn{2}{c|}{No Local Scale} & \multicolumn{2}{c}{Our Method}       \\ \cline{3-10} 
		&          & When        & Whether          & When         & Whether           & When            & Whether           & When           & Whether              \\ \hline
		\multirow{3}{*}{LSTM}        & Single   & 0.808       & 0.879            & 0.822        & 0.889             & 0.820           & 0.888             & 0.818          & 0.926                \\ 
		& Transfer & 0.237       & 0.769            & 0.453        & 0.763             & 0.494           & 0.801             & 0.491          & 0.798                \\ 
		& Finetune & 0.796       & 0.890            & 0.818        & 0.881             & 0.788           & 0.860             & {\ul 0.837}    & {\ul \textbf{0.912}} \\ \hline
		\multirow{3}{*}{Transformer} & Single   & 0.776       & 0.872            & 0.740        & 0.874             & 0.741           & 0.869             & 0.740          & 0.877                \\ 
		& Transfer & 0.300       & 0.741            & 0.358        & 0.736             &    0.437             &     0.782              & 0.433          & 0.777                \\ 
		& Finetune & 0.723       & 0.748            & 0.728        &  0.865       &      0.710           &        0.840           & {\ul 0.751}    & {\ul 0.882}              \\ \hline
		\multirow{3}{*}{Mixed}       & Single   & 0.841       & 0.901            & 0.847        & 0.885             & 0.837           & 0.895             & 0.848          & {\ul 0.911}               \\ 
		& Transfer & 0.242       & 0.781            & 0.253        & 0.835             & 0.546           & 0.805             & 0.542          & 0.812                \\ 
		& Finetune & 0.852       & 0.901      & 0.833        & 0.877             & 0.847           & 0.892             & {\ul \textbf{0.878} } & 0.909                \\ \hline
	\end{tabular}
    \caption{Ablation Study on FT dataset. \emph{Zero Norm} uses the common zero-score normalization instead of the proposed rotation normalization (RN); \emph{No Rotation} does not use the proposed Orientation Invariance (OI); \emph{No Local Scale} does not use the proposed Coordinate Invariance (CI). \textbf{Bold} represents the best under the current metric among all and {\ul underline} represents the best under the current metric with the current model.}
    \label{tab:ablation}
\end{table*}

\begin{table*}[!h]
	\centering
	\scriptsize
	\begin{tabular}{c|cc|cc|cc|cc}
		\hline
		\multirow{2}{*}{Method} & \multicolumn{2}{c|}{Zero Norm} & \multicolumn{2}{c|}{No Rotation} & \multicolumn{2}{c|}{No Local Scale} & \multicolumn{2}{c}{Our Method} \\ \cline{2-9} 
		& When          & Whether        & When           & Whether         & When            & Whether           & When          & Whether         \\ \hline
		LSTM                    & 0.835         & 0.886          & 0.851          & 0.889           & 0.858           & 0.887             & {\ul 0.878}         & {\ul 0.891}           \\ \hline
		Transformer             & 0.800         & 0.879          & 0.785          & 0.871           & 0.799           & 0.872             & {\ul 0.812}         & {\ul 0.887}           \\ \hline
		Mixed                   & 0.856         & 0.898          & 0.865          & 0.895           & 0.878           & 0.885             & {\ul \textbf{0.882}}         & {\ul \textbf{0.910}}           \\ \hline
	\end{tabular}
	
 \caption{Ablation Study on all scenarios. \emph{Zero Norm} uses the common zero-score normalization instead of the proposed rotation normalization (RN); \emph{No Rotation} does not use the proposed Orientation Invariance (OI); \emph{No Local Scale} does not use the proposed Coordinate Invariance (CI). \textbf{Bold} represents the best under the current metric among all and {\ul underline} represents the best under the current metric with the current model.}
    \label{tab:ablation-full}
	\end{table*}

We also did an ablation study to show the effectiveness of our proposed three data preprocessing methods. First, we did an ablation study on the FT dataset (which has the largest number of samples). We tested the performance under three different training settings: 1) \emph{Single} means training and testing with the only one scenario, 2) \emph{Transfer} means a training with all the other scenarios and test on different one, and 3) \emph{Fine-tune} means training with all the other scenarios and then fine-tuning with the current one. We show the results under three different Spatial-Temporal Blocks: 1) only LSTM layers, 2) only Transformer layers, and 3) the proposed mixed LSTM and Transformer. The results are in Tab. \ref{tab:ablation}.

From Tab.~\ref{tab:ablation}, we can conclude that: 1. Under the Fine-tune setting which has the best performance, all three data pre-processing methods could improve the model's performance. 2. Under the Single Setting, not all the pre-processing methods help improve the performance. We think it may be because removing one of the pre-processing method could make the model much easier to overfit the scenarios. However, the generalization ability is poor. To further verify this hypothesis, we trained and tested with all the scenarios to see how the data pre-processing methods could influence the results when the train and test data both contain diverse driving scenarios. The results is in Tab. \ref{tab:ablation-full}. We can see in Tab. \ref{tab:ablation-full}, removing any of the three data pre-processing methods would significantly reduce the performance in terms of generalizability.

\section{Hyper-Parameter Setting}
In the experiment, we use GELU as the activatation function, AdamW as optimizer, and warm-up learning rate schedule. We set weight\_decay as 1e-7, learning rate as 3e-6, epoch as 250,  dropout as 0.01, ratio of warm-up as 0.01, and gradient-clip as 10. As for model, we use hidden-dimension as 384, head-num as 16. All module has 2 Spatial-Temporal Blocks (ST-B). For the trajectory prediction, we predict 5 future steps. The weights of each loss terms:  Supervised Whether Loss - 0.233, Supervised When Loss - 0.233,  Trajectory Prediction Loss - 0.007, Prior Loss - 0.233, Uncertainty Loss - 0.007, Rotation-invaraince Loss - 0.023.

\end{document}